\documentclass[10pt,conference]{IEEEtran}

\usepackage{cite}
\usepackage{hyperref}

\ifCLASSINFOpdf
   \usepackage[pdftex]{graphicx}
   \graphicspath{{figs/}}
   \DeclareGraphicsExtensions{.pdf,.jpeg,.png}
\else
   \usepackage[dvips]{graphicx}
   \graphicspath{{../figs/}}
   \DeclareGraphicsExtensions{.eps}
\fi
\usepackage[cmex10]{amsmath}
\usepackage{amsthm}

\usepackage{algorithmic}
\usepackage{array}
\ifCLASSOPTIONcompsoc
  \usepackage[caption=false,font=normalsize,labelfont=sf,textfont=sf]{subfig}
\else
  \usepackage[caption=false,font=footnotesize]{subfig}
\fi
\usepackage{url}
\usepackage{todonotes}

\usepackage[switch]{lineno}

\newif\iffinal
\finalfalse
\finaltrue
\newcommand{\cmtid}{163}
\iffinal
\else
\fi

\IEEEoverridecommandlockouts
\IEEEpubid{%
  \begin{minipage}{\textwidth}
    \scriptsize
    \vspace{2cm}
    \copyright 2025 IEEE. Personal use of this material is permitted. Permission from IEEE must be obtained for all other uses, in any current or future media, including reprinting/republishing this material for advertising or promotional purposes, creating new collective works, for resale or redistribution to servers or lists, or reuse of any copyrighted component of this work in other works. This is the author's accepted manuscript of an article that will appear in the 2025 38th Conference on Graphics, Patterns and Images (SIBGRAPI). The final version is available at: \url{https://doi.org/10.1109/SIBGRAPI67909.2025.11223386}
  \end{minipage}
}

\begin{document}
%
% paper title
% Titles are generally capitalized except for words such as a, an, and, as,
% at, but, by, for, in, nor, of, on, or, the, to and up, which are usually
% not capitalized unless they are the first or last word of the title.
% Linebreaks \\ can be used within to get better formatting as desired.
% Do not put math or special symbols in the title.
\title{
Advancing Annotat3D with Harpia: A CUDA-Accelerated Library For Large-Scale Volumetric Data Segmentation
}

% author names and affiliations
% use a multiple column layout for up to two different
% affiliations

\iffinal

% author names and affiliations
% use a multiple column layout for up to three different
% affiliations
\author{
\IEEEauthorblockN{Camila Machado de Araujo\IEEEauthorrefmark{1},
Egon P. B. S. Borges\IEEEauthorrefmark{1},
Ricardo Marcelo Canteiro Grangeiro\IEEEauthorrefmark{1}, and
Allan Pinto\IEEEauthorrefmark{1}}
\IEEEauthorblockA{\IEEEauthorrefmark{1}Brazilian Synchrotron Light Laboratory (LNLS)\\
Brazilian Center for Research in Energy and Materials (CNPEM)\\
Campinas, Sao Paulo, Brazil\\ Emails:
\{camila.araujo, egon.borges, ricardo.grangeiro, allan.pinto\}@lnls.br}}

% conference papers do not typically use \thanks and this command
% is locked out in conference mode. If really needed, such as for
% the acknowledgment of grants, issue a \IEEEoverridecommandlockouts
% after \documentclass

% for over three affiliations, or if they all won't fit within the width
% of the page, use this alternative format:
% 
% \author{\IEEEauthorblockN{Michael Shell\IEEEauthorrefmark{1},
% Homer Simpson\IEEEauthorrefmark{1},
% James Kirk\IEEEauthorrefmark{1}, \\
% Montgomery Scott\IEEEauthorrefmark{1} and
% Eldon Tyrell\IEEEauthorrefmark{1}}
% \IEEEauthorblockA{\IEEEauthorrefmark{1}School of Electrical and Computer Engineering\\
% Georgia Institute of Technology,
% Atlanta, Georgia 30332--0250\\ Email: see http://www.michaelshell.org/contact.html}
% }

\else
  \author{SIBGRAPI Paper ID: \cmtid \\ }
  \linenumbers
\fi

\newcommand{\anon}[1]{#1}

% make the title area
\maketitle

\IEEEpubidadjcol

% \begin{itemize}
%     \item motivation/context
%     \item GAP/Briefly state the problem (interactive segmentation of large 3D tomographic volumes)
%     \item purpose of work
%     \item methodology
%     \item Main Results/contributions (GPU-accelerated algorithms, interactive tools, scalable web interface, benchmarking, real use in scientific environments).
%     \item conclusion
% \end{itemize}

% As a general rule, do not put math, special symbols or citations
% in the abstract
\begin{abstract}

High-resolution volumetric imaging techniques, such as X-ray tomography and advanced microscopy, generate increasingly large datasets that challenge existing tools for efficient processing, segmentation, and interactive exploration. This work introduces new capabilities to Annotat3D through Harpia, a new CUDA-based processing library designed to support scalable, interactive segmentation workflows for large 3D datasets in high-performance computing (HPC) and remote-access environments. Harpia features strict memory control, native chunked execution, and a suite of GPU-accelerated filtering, annotation, and quantification tools, enabling reliable operation on datasets exceeding single-GPU memory capacity. Experimental results demonstrate significant improvements in processing speed, memory efficiency, and scalability compared to widely used frameworks such as NVIDIA cuCIM and scikit-image. The system's interactive, human-in-the-loop interface, combined with efficient GPU resource management, makes it particularly suitable for collaborative scientific imaging workflows in shared HPC infrastructures.

\end{abstract}

% CUDA acceleration, volumetric data segmentation, high-performance computing (HPC), synchrotron imaging, GPU memory management, chunked processing architecture, interactive segmentation, image processing, image analysis.

% For peerreview papers, this IEEEtran command inserts a page break and
% creates the second title. It will be ignored for other modes.
\IEEEpeerreviewmaketitle

\section{Introduction and Motivation}

Advances in high-resolution X-ray tomography have revolutionized research across porous media, materials science, biology, and synchrotron applications. Particularly when deployed at synchrotron light sources and advanced microscopy facilities, this imaging technique provides unprecedented insights into internal structures at micro and nanoscales~\cite{anon2025a}. Nevertheless, in some experimental settings, massive volumetric datasets, reaching terabytes of data, can be produced in few hours in a single experiment, which poses significant technical challenges for efficient processing, visualization, and segmentation~\cite{Wang2018, anon2025a}.

Interactive, expert-guided segmentation remains a crucial step in extracting quantitative information from these complex volumes. However, existing tools often struggle to meet the combined demands of large-scale data handling, low-latency performance, and integration into high-performance computing (HPC) environments. This limitation is particularly acute in facilities that rely on shared, centralized compute infrastructure, where remote access, multi-user support, and scalable resource management are essential.

Several widely used software platforms illustrate these limitations. Avizo~\cite{ThermoFisher_Avizo} offers powerful analysis capabilities but relies on complex, proprietary workflows and requires expensive licenses or extensions to efficiently process large volumetric datasets, limiting accessibility and integration into open scientific environments. Fiji~\cite{Schindelin2015}, while popular for 2D and basic 3D analysis, lacks native support for scalable Graphics Processing Unit (GPU) acceleration and multi-user deployment on HPC clusters. Performance may degrade significantly with large 3D stacks, limiting its applicability to large datasets. 3D Slicer~\cite{Kikinis2014}, designed primarily for medical imaging, provides GPU-accelerated rendering for visualization tasks but relies largely on CPU-based processing for segmentation and filtering. As a result, it exhibits poor responsiveness and limited memory management when applied to large scientific datasets, restricting its scalability in HPC environments. Finally, DragonFly~\cite{Gendron2017} provides advanced segmentation features but remains a closed, licensed software with limited options for integration into flexible, web-based, HPC-oriented workflows.

Recent research has explored machine learning, GPU acceleration, and web-based interfaces to overcome these barriers. Despite recent advances, there remains a scarcity of open-source, modular tools that offer an integrated, scalable solution specifically designed to address the demands of scientific users working with large-scale volumetric datasets. 

This work introduces a series of advancements in \anon{Annotat3D}~\cite{anon2025b}.\footnote{\url{https://github.com/cnpem/annotat3d}} The existing platform already supports interactive segmentation with lightweight ML-assisted tools, CUDA-accelerated superpixel over-segmentation and feature extraction, integration with scikit-image filters, and a web-based front-end supporting simple brush annotation. In this work, we present Harpia\footnote{\url{https://github.com/cnpem/harpia}} a novel CUDA C++ backend, designed to address the increasing computational and data demands of large-scale image segmentation. The main technical contributions include: 

\begin{enumerate}
    \item  \textbf{Scalable Chunked-based Processing Architecture:} A context-aware, chunked execution model that enables efficient processing of datasets far exceeding single-GPU memory capacity. This design ensures scalability in line with the increasing data volumes generated by modern experimental facilities;

    \item  \textbf{Efficient GPU Resource Management:} A lightweight execution approach that releases memory and computation resources upon task completion, supporting concurrent multi-user workflows in HPC environments;

    \item  \textbf{Label Editing Module (3D):} A new label editing workflow with fully CUDA-accelerated 3D morphological operations, thresholding methods, and island removal. A C++ 2.5D Watershed implementation further supports precise segmentation refinement;

    \item  \textbf{High-Performance Filtering Suite:} CUDA-accelerated implementations of pre- and post-processing filters, including Unsharp Mask, Anisotropic Diffusion, Median, and Non-Local Means;

    \item  \textbf{Quantification Module:} Tools for post-segmentation analysis, including CUDA-accelerated volume, area, perimeter, and fraction computation, as well as connected components labeling and Euclidean Distance Transform (via OpenMP); and
    
    \item  \textbf{Extended Annotation Tools:} New intuitive tools (e.g., Magic Wand, Lasso) and accelerated 2D algorithms, including CUDA-accelerated implementations of active contours (Snakes), morphological and thresholding operations. 
\end{enumerate}

The remainder of this paper is structured as follows: Section~\ref{sec:related_work} reviews related work across the relevant domains. Section~\ref{sec:methodology} presents the architecture of \anon{Annotat3D}, including its algorithmic foundations, GPU-accelerated methods, and the interactive, web-based interface. Section~\ref{sec:experiments} presents performance benchmarks comparing \anon{Annotat3D} with other established frameworks. % and application-driven use case demonstrations. 
Finally, Section~\ref{sec:conclusion} summarizes the main findings and discusses directions for future work.

\section{Related Work}
\label{sec:related_work}

The proposed work is situated at the intersection of volumetric image segmentation, GPU-accelerated computing, and web-based scientific applications. To contextualize our contributions, this section provides an overview of representative tools and research efforts from key domains, including medical imaging, porous media analysis, and synchrotron applications. Despite their specificity, these fields share common challenges such as the management of large volumetric datasets, low-contrast images, domain-specific segmentation requirements, and the necessity for scalable, and interactive workflows that can efficiently operate within HPC environments.

Several studies from the aforementioned domains are reviewed, emphasizing their strengths and limitations in performance, interactivity, and scalability. This comparative analysis establishes the motivation behind \anon{Annotat3D's} architecture and capabilities, showing how our system advances the state of the art in large-scale volume segmentation by integrating GPU-accelerated processing, chunk-based processing, and human-in-the-loop segmentation within a modern web interface. This design enables low-latency, scalability and remote interaction, which are critical requirements for data-intensive scientific applications operating in HPC environments.

\subsection{Medical Imaging}
Medical imaging remains a central domain of research for volumetric data segmentation, encompassing a broad range of modalities, including magnetic resonance imaging (MRI), computed tomography (CT), positron emission tomography (PET), ultrasound, microscopy, as well as emerging hyperspectral and multimodal imaging techniques. Recent advances in this field have primarily focused on segmentation tasks related to oncology~\cite{Florimbi20208485}%Zhang2023433,Lin20236548,Liu2025
, organ segmentation for personalized treatment~\cite{Geroski2023, Himthani2022, Khokhariya2024, Aly2025, Kumar2025, Qian2024}, and neuroimaging applications targeting neurological and cerebrovascular disorders~\cite{Valevicius2023, Chen2025, Gao2023}. %Ak2022792

In CT, deep learning models such as residual convolutional sparse coding networks for low-dose imaging~\cite{Liu2021} and UNet-based methods for organ segmentation~\cite{Khokhariya2024} have demonstrated significant improvements in segmentation accuracy, noise reduction, and artifacts suppression. In the context of ultrasound imaging, recent advances have addressed challenges related to limited annotations and low-contrast images through weakly supervised, multi-branch frameworks~\cite{Aly2025} and contrastive pretraining strategies for left ventricle segmentation~\cite{Qian2024}. For MRI, Transformer-based architectures have been shown to outperform traditional convolutional neural networks (CNNs) in the segmentation of white matter hyperintensities, owing to their superior capacity for global context modeling~\cite{Chen2025}.

The practical usage of these methods depends on scalable, robust platforms capable of supporting data-intensive processing. Some examples include web-based systems such as DPLab, which facilitates 3D pathology reconstruction~\cite{Shen2022}, and HPC pipelines designed for the processing of extensive functional MRI datasets~\cite{Gao2023, Valevicius2023}. Furthermore, FAIR-compliant, HPC-integrated platforms such as LABKIT~\cite{Arzt2022}, SIMPLI~\cite{Bortolomeazzi2022}, and BIOMERO~\cite{Luik2024} have played a key role in democratizing access to advanced bioimage analysis, making these capabilities available to a broad and diverse scientific community.

% Bioinformatics research increasingly benefits from advanced imaging and data processing pipelines, particularly cryo-electron microscopy (cryo-EM) workflows~\cite{Shell2024147} and high-throughput platforms for structural biology and digital pathology analysis~\cite{Luik2024, Shen2022}. Recent efforts have also emphasized user-friendly, FAIR-compliant software solutions that facilitate large-scale bioimage analysis for a broad scientific community~\cite{Luik2024, Arzt2022, Li20231691}.

% Overall, recent developments in medical imaging and bioinformatics are converging towards the establishment of AI-driven, HPC-enabled analytical ecosystems. These platforms seamlessly integrate advanced computational methods, including self-supervised and weakly supervised learning, with scalable cloud infrastructures and FAIR-compliant software frameworks. This convergence enables accurate, efficient, and reproducible segmentation and analysis of large-scale volumetric datasets, ultimately supporting clinical, biological, and structural research at unprecedented levels of scale and accessibility.

\subsection{Porous Media and Synchrotron Applications}
The segmentation of porous materials and synchrotron-generated volumetric datasets presents distinct scientific and computational challenges. These challenges are primarily associated with the inherently low contrast of X-ray computed tomography ($\mu$CT) images, the considerable size of the datasets that often exceeding billions of voxels, therefore demanding HPC infrastructures while accommodating domain-specific expert intervention.

Within porous media research, precise and reproducible segmentation is a prerequisite for the quantitative characterization of pore structures, enabling the simulation of transport phenomena at both micro- and nanoscale. Richert et al.~\cite{Richert2020} provide a comprehensive review of nanoporous metals, emphasizing that structural characterization based on advanced 3D imaging techniques as nanotomography combined with robust image processing and segmentation tools is essential for establishing correlations between microstructural features and mechanical, electrochemical, and functional properties. Such analyses critically depend on scalable and HPC-oriented segmentation workflows capable of efficiently processing complex, high-resolution volumetric datasets.

% Similarly, An et al.~\cite{An2020} integrate GPU-accelerated image segmentation directly into pore-scale two-phase flow simulations. Their method combines a volumetric lattice Boltzmann approach with level-set-based segmentation, establishing a seamless pipeline from image acquisition to computational fluid dynamics (CFD) simulation. This integration of image analysis within physics-based simulations demonstrates the critical role of segmentation in predictive modeling of porous media, particularly in fields such as hydrogeology and reservoir engineering.

% In the agricultural domain, image-based segmentation also plays a vital role. López-Martínez et al.~\cite{Lopez-Martinez2023} leverage HPC clusters and deep learning (CNNs) to perform weed classification from agricultural imagery, demonstrating how large-scale, GPU-enabled image analysis can support precision agriculture, which often shares imaging challenges with porous media studies (e.g., high-resolution, variable contrast).

Synchrotron light sources further intensify these challenges by producing increasingly complex, high-resolution volumetric datasets, which are essential for revealing the internal structures of materials across a broad spectrum of scientific disciplines. McClure et al.~\cite{McClure2021226} present an integrated experimental workflow for synchrotron-based X-ray $\mu$CT that combines deep learning-based denoising, conventional segmentation techniques, and quantitative analysis, all executed on Summit, hosted at Oak Ridge National Laboratory, USA. Although the segmentation stage of their workflow primarily employs traditional algorithms, the study highlights the critical need for scalable, HPC-oriented workflows that seamlessly incorporate artificial intelligence to meet the demands of large-scale, data-intensive experimental environments.

The pressing need for scalable software tools capable of managing the data rates and complexity associated with next-generation synchrotron facilities is underscored by Li et al.~\cite{Li2023} that emphasize the increasing importance of unified software frameworks that seamlessly integrate AI-assisted image processing, such as segmentation and denoising, directly into experimental control and data acquisition pipelines. %Such integrated approaches are indispensable for fully exploiting the capabilities of modern synchrotron beamlines and effectively addressing the data-intensive challenges posed by multidimensional and multimodal experiments.

\subsection{Final Remarks}

As we discussed in this section, common challenges persist in the segmentation and analysis of large-scale volumetric datasets, across diverse domains, ranging from medical imaging to porous media analysis, and synchrotron applications. These include the management of increasingly complex, high-resolution images, the need for domain-specific yet scalable segmentation workflows, and the integration of advanced computational methods within interactive, human-in-the-loop environments.

While recent efforts have made significant progress through the application of machine learning, AI-assisted methods, and HPC infrastructures, existing solutions often exhibit critical limitations. Many tools rely on proprietary software, lack seamless integration with HPC clusters, or emphasize batch processing over interactive, iterative workflows essential for expert-driven analyses.

\section{Materials and Methods}
\label{sec:methodology}

This section details the architecture, computational strategies, and interactive components of \anon{Annotat3D}, a GPU-accelerated, web-based platform designed to address the challenges of interactive segmentation for large-scale volumetric datasets in HPC environments. Building upon the limitations identified in existing tools, \anon{Annotat3D} advances the state of the art by integrating several key elements into a unified and scalable solution: (i) a browser-accessible interface that eliminates the need for local software installation while enabling responsive interaction with remote HPC resources; (ii) a suite of CUDA-accelerated image processing algorithms designed to enhance segmentation accuracy, considering efficiency aspects in memory utilization; (iii) a chunked, context-aware processing strategy that supports datasets larger than single-GPU memory capacity, common in synchrotron-based X-ray imaging; (iv) a comprehensive set of segmentation-aiding tools, including superpixel over-segmentation, watershed algorithms, and intuitive graphical interaction tools, and (v) machine learning methods to scale segmentation across complete 3D volumes.
Together, these components provide an integrated framework for data-intensive scientific applications.

%Special consideration has been given to 
\anon{Annotat3D} is tailored to the requirements of advanced imaging facilities, such as synchrotron light sources, which routinely generate massive, high-resolution volumetric datasets ~\cite{anon2025a}. Its browser-based interface facilitates broad accessibility for geographically distributed research teams, a critical requirement in synchrotron-based experimental workflows.

At the core of the system lies Harpia (High Algorithmic Performance for Image Analysis), a CUDA-accelerated library for image quality enhancement, structure delineation, and segmentation refinement in complex, low-contrast samples typical of porous media, biology, and materials science. Harpia implements a chunked, context-aware strategy that partitions volumes into sub-volumes while maintaining consistency across chunk boundaries—a prerequisite for accurate and reproducible segmentation in large-scale tomography experiments~\cite{anon2025a}.

Complementing these computational strategies, \anon{Annotat3D} provides a comprehensive set of interactive tools for human-in-the-loop segmentation. Superpixel over-segmentation, watershed methods, and graphical interfaces enable low-latency, high-fidelity workflows that integrate expert knowledge directly into the analysis process.

%Finally, \anon{Annotat3D} provides new gpu-accelerated implementations of established feature extraction and segmentation refinement techniques, supported by machine learning methods, which leverage user interaction within a human-in-the-loop strategy to iteratively enhance segmentation accuracy and adapt results to domain-specific requirements. 

Finally, the platform extends functionality with new GPU-accelerated implementations of feature extraction and segmentation refinement techniques, coupled with machine learning models that iteratively adapt results to domain-specific requirements. This combination of scalable computation, interactive refinement, and learning-based generalization makes \anon{Annotat3D} a powerful tool for modern, data-intensive scientific imaging.

%Figure~\ref{fig:architecture_overview} provides a schematic overview of the system architecture, illustrating the main components, including the web-based frontend, GPU-accelerated backend, and the data processing pipeline that enables scalable, interactive segmentation workflows.
%
% \begin{figure*}
% 	\centering
% 	\includegraphics[width=0.5\textwidth]{figs/annotat3Dweb_arch.pdf}
% 	\caption{\anon{Annotat3D} architecture overview.}
% 	\label{fig:architecture_overview}
% \end{figure*}

\subsection{System Architecture and Processing Pipeline}

\anon{Annotat3D} adopts a modular client–server architecture tailored for interactive image segmentation in remote HPC environments. The system operates entirely through the web: a browser-based frontend communicates with the backend via a RESTful API over standard HTTP, enabling secure, low-latency interaction with GPU-accelerated infrastructure. The backend, implemented in CUDA C++, exploits modern GPUs and HPC clusters to perform large-scale image processing and segmentation. Software architecture is illustrated in Fig.~\ref{fig:frontend}.

%\anon{Annotat3D} adopts a modular client-server architecture designed to support interactive segmentation in remote  HPC environments. The system is optimized for web-based operation, with the frontend and backend communicating through a RESTful API over standard HTTP protocols. This architecture enables secure, low-latency interaction between the browser-accessible user interface and the underlying GPU-accelerated computational infrastructure. The backend, implemented in CUDA C++, efficiently leverages the computational capabilities of modern GPUs and HPC clusters to perform large-scale data processing and segmentation tasks.

To bridge the CUDA C++ backend with the web interface, we developed Harpia, a modular library for high-performance image analysis. Harpia is exposed through a Python interface and delivered as a RESTful API, ensuring language-agnostic access from the \anon{Annotat3D} frontend. This design gives users web-based access to core computational tasks—including filtering, segmentation, and morphological operations—while preserving responsiveness and scalability.

%To enable seamless integration between the CUDA C++ backend and the web-based frontend, we developed Harpia, a highly modular software library designed to deliver high-performance image analysis capabilities. Harpia encapsulates low-level image processing, segmentation, and feature extraction algorithms implemented in CUDA C++, providing a robust and scalable foundation for volumetric data analysis.  A Python interface wraps Harpia functionalities and exposes them as a RESTful API, enabling efficient, language-agnostic communication with the \anon{Annotat3D} frontend. The exposed RESTful API provides access to core functionalities, allowing users to perform computationally intensive tasks including filtering, segmentation, and morphological operations, directly from the web interface, while maintaining responsiveness and scalability.

%The backend is implemented using a combination of Python and CUDA C++, ensuring seamless integration with CUDA-accelerated processing routines. The exposed RESTful API provides efficient access to core functionalities, allowing users to perform computationally intensive tasks including filtering, segmentation, and morphological operations, directly from the web interface, while maintaining responsiveness and scalability. A fundamental aspect of the platform is its chunked, context-aware processing strategy, which enables the efficient handling of volumetric datasets that significantly exceed the memory capacity of a single GPU.  

The platform runs efficiently across HPC environments, from GPU-equipped workstations to large supercomputers. Task scheduling mechanisms dynamically allocate and release GPU resources after each processing step, enabling concurrent multi-user workflows and minimizing resource contention. 
Furthermore, the modular design facilitates integration with HPC resource managers and job schedulers, simplifying deployment on shared infrastructures common in synchrotron facilities, research centers, and cloud platforms.
%The platform is designed to operate seamlessly within HPC environments, including GPU-equipped workstations, compute clusters, and large-scale supercomputers. Task scheduling mechanisms ensure that GPU resources are dynamically allocated and released after each processing step, supporting efficient multi-user operation and preventing resource contention. 

%Furthermore, the modular design allows easy integration with HPC resource managers and job schedulers, facilitating deployment on shared infrastructures typical of synchrotron facilities, research centers, and cloud platforms. 

%\subsection{Web-Based Frontend and Human-in-the-Loop Workflow}
\subsection{Web-Based Frontend and Extended Annotation Tools}

%Extended Annotation Tools

\begin{figure}
    \centering
    \includegraphics[width=\linewidth]{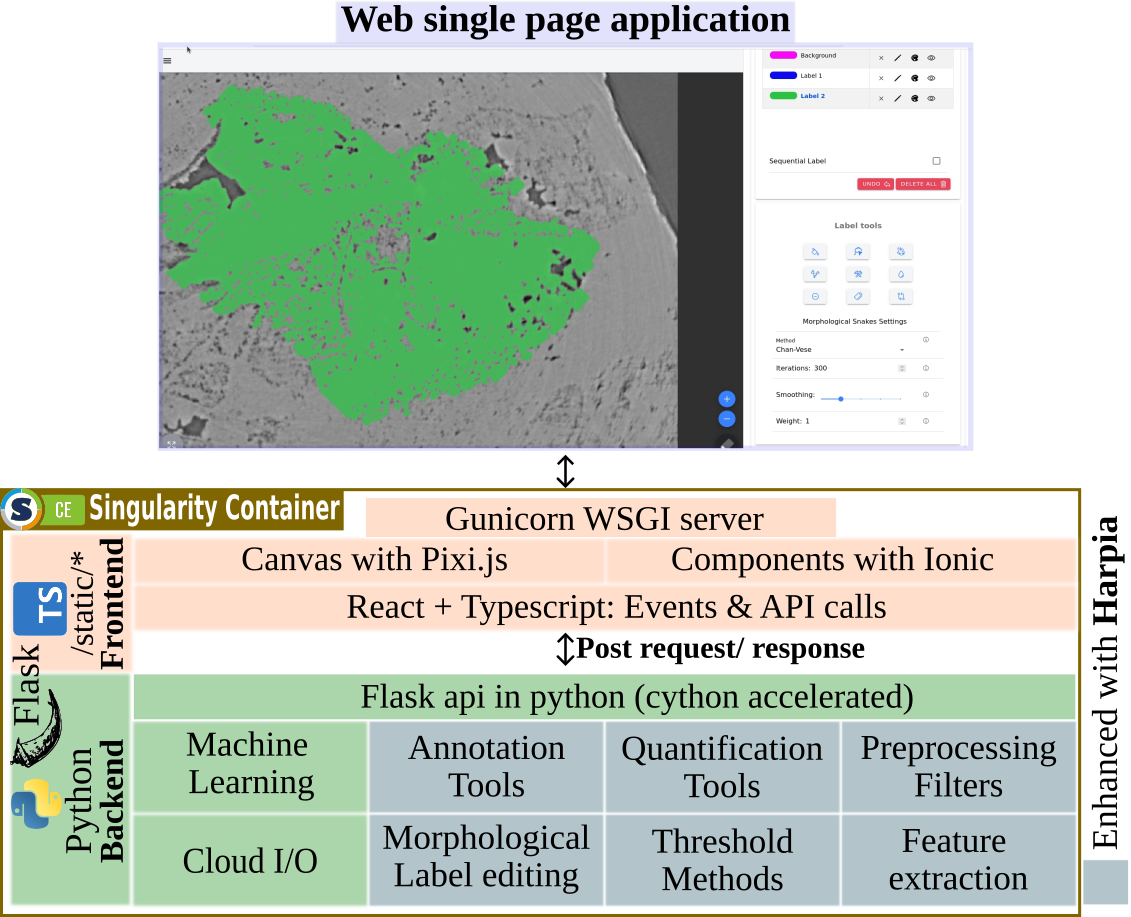}
    \caption{Overall architecture of the \anon{Annotat3D} web application.}
    \label{fig:frontend}
\end{figure}

The front end of \anon{Annotat3D} is a lightweight browser-accessible application built using modern web technologies (e.g., Pixi.js, TypeScript, and React) (Fig.~\ref{fig:frontend}). This design eliminates the need for specialized client installations, enabling users to interactively explore, segment, and annotate volumetric datasets from standard web browsers.
A key feature of the platform is its support for \textit{human-in-the-loop} segmentation workflows. In this work, we extended annotation capabilities with new intuitive tools:

\begin{itemize}
    \item \textbf{Magic Wand:} Semi-automatic region growing based on intensity similarity.
    \item \textbf{Lasso Tool:} Freehand selection for manual delineation of regions of interest (ROIs).
    \item \textbf{GPU-accelerated Morphological Snakes Tool:} Interactive contour evolution guided by user inputs and local image features.
\end{itemize}

These tools allow users to iteratively refine segmentation outputs by combining manual inputs with automated algorithms, thereby improving accuracy in challenging scenarios, such as low-contrast regions typical in porous media or biological samples.

%\subsection{Preprocessing and Segmentation-Aiding Algorithms}
\subsection{Harpia Library}

Harpia encapsulates low-level image analysis algorithms implemented in CUDA C++, providing a robust and scalable foundation for volumetric data analysis. The suite of accelerated algorithms, include:

\begin{itemize}
    \item \textbf{Label Editing Module:} CUDA C++ implementations of morphological operations, including Erosion, Dilation, Opening, Closing, Geodesic Reconstruction, Smoothing, and Fill Holes; threshold algorithms, including Niblack, Savoula, Mean, Median, Gaussian, Otsu; and a C++ implementation of 2.5D Watershed.
    \item \textbf{Filtering and Feature Extraction Suite:}  CUDA-accelerated implementations of Unsharp Masking, Sobel and Prewitt for edge enhancement; smoothing and denoising filters such as Mean, Median, Gaussian and Non-Local-Means; Anisotropic Diffusion for noise reduction while preserving structural boundaries; and texture feature extraction with Local Binary Patterns (LBP) and Hessian.
    \item \textbf{Quantification Module:} A suite of CUDA implementations including quantification of area, volume, perimeter, fraction, connected components, and as Open MPI implementation of Euclidean Distance Transform (EDT).
    \item \textbf{Accelerated annotations:} New accelerated brushes include a CUDA implementation of active contours. All 3D morphological and threshold-based label editing tools are also available, with support for fine annotation editing in 2D.
    
\end{itemize}

The library enables users to prepare datasets for segmentation in an interactive manner, ensuring optimal visual conditions for subsequent analysis. It also allows fine annotations editing, post-processing, and refinement of segmented volumes. Harpia is part of an ongoing initiative at \anon{Sirius} to advance image analysis capabilities for \anon{synchrotron-based research}. Its development emphasizes scalable, GPU-accelerated algorithms with a strong focus on parallel programming and performance optimization. A fundamental aspect of the library is its chunked, context-aware processing strategy, which enables the efficient handling of volumetric datasets that exceed the memory capacity of a single GPU.
%These operations are GPU-accelerated to maintain interactivity, even on large 3D datasets. %Finally, superpixels, which group spatially coherent voxels based on intensity and texture similarity, play a crucial role in facilitating rapid high-quality segmentation within \anon{Annotat3D}~\cite{Machairas2015}. Superpixels serve two primary purposes:
%
%\begin{enumerate}
%    \item They provide an over-segmentation of the dataset, reducing the number of elements that require manual inspection or annotation;
%    \item They act as a basis for feature extraction, enabling downstream machine learning or clustering methods to operate on compact, meaningful regions rather than individual voxels.
%\end{enumerate}
% Users can interactively adjust superpixel granularity and utilize them as inputs for the Magic Wand or Snakes tools, combining manual and algorithmic segmentation strategies within a human-in-the-loop framework.
\subsection{Chunked-based Processing Architecture}
%\subsection{Scalability and HPC Integration}
                         
To operate on images far exceeding single-GPU memory, a GPU-aware volume partitioning strategy is employed. Input volumes are subdivided into overlapping, context-preserving blocks (chunks) for independent GPU processing. The inclusion of overlapping regions corrects boundary artifacts, preserves structural continuity across chunk edges, and ensures the seamless reconstruction of the final segmented volume. The execution model follows three steps: 

\begin{enumerate}
    \item \textbf{GPU Resource Profiling: }Free GPU memory is queried, and a fixed fraction (set at library level) is reserved for chunking. Dynamic tuning is a target for future work. 
    \item \textbf{Chunk Size Estimation:} The library computes the maximum number of Z-slices per chunk, including padding to maintain continuity. 
    \item \textbf{Independent Chunk Execution:} The volume is split along the Z-axis and processed chunk by chunk. Current implementation uses a single GPU, but multi-GPU support is planned via round-robin scheduling without inter-GPU communication. 
\end{enumerate}

This strategy ensures predictable memory usage, data locality, and scalability. Future work includes dynamic tuning and full multi-GPU support for large-scale HPC deployments. 

\section{Experimental Results and Discussion}
\label{sec:experiments}

This section presents the experimental evaluation of \anon{Annotat3D}, focusing on efficiency, scalability, and applicability in comparison with two established baselines: scikit-image (CPU-based) and NVIDIA cuCIM (GPU-accelerated).

\subsection{Experimental Setup}

All experiments were conducted on a dedicated compute node equipped with an NVIDIA L40S GPU. To ensure statistical robustness, each measurement was averaged over 30 independent runs, from which we compute the following metrics:  (i) Memory footprint, including residual and peak memory consumption, in Gibibytes (GiB); and (ii) Execution time, in seconds (s).

The evaluation employed a real-word volumetric dataset acquired in the \anon{MOGNO Beamline at Sirius}, with $2048 \times 2052 \times 2052$. To assess scalability, we sliced the volume to simulate dataset sizes ranged from $1$~GiB to $32$~GiB, approximately.

\subsection{Performance Comparison with Baselines}

\begin{figure}
    \centering
    \includegraphics[width=1\linewidth]{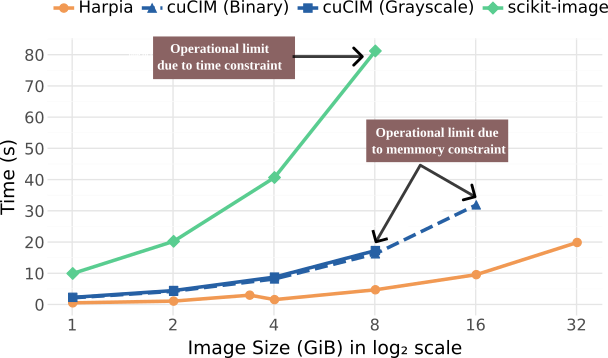}
    \caption{Performance evaluation in terms of execution time.}
    \label{fig:res-execution-time}
\end{figure}

Figure~\ref{fig:res-execution-time} shows execution time as a function of dataset size. The CPU-based scikit-image degraded rapidly, becoming impractical beyond $8$~GiB. The GPU-accelerated cuCIM outperformed the CPU baseline but failed at the $32$~GiB scale. In contrast, \anon{Annotat3D} consistently achieved the lowest execution times and successfully processed the largest datasets.

\begin{figure}
    \centering
    \subfloat[Residual Memory]{\includegraphics[width=0.4\linewidth]{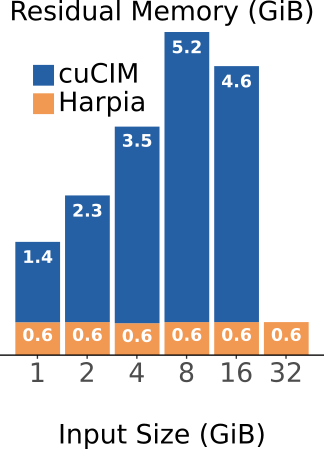}
    \label{fig:res-residual-mem}}
    \hfil
    \subfloat[Maximum Memory]{\includegraphics[width=0.4\linewidth]{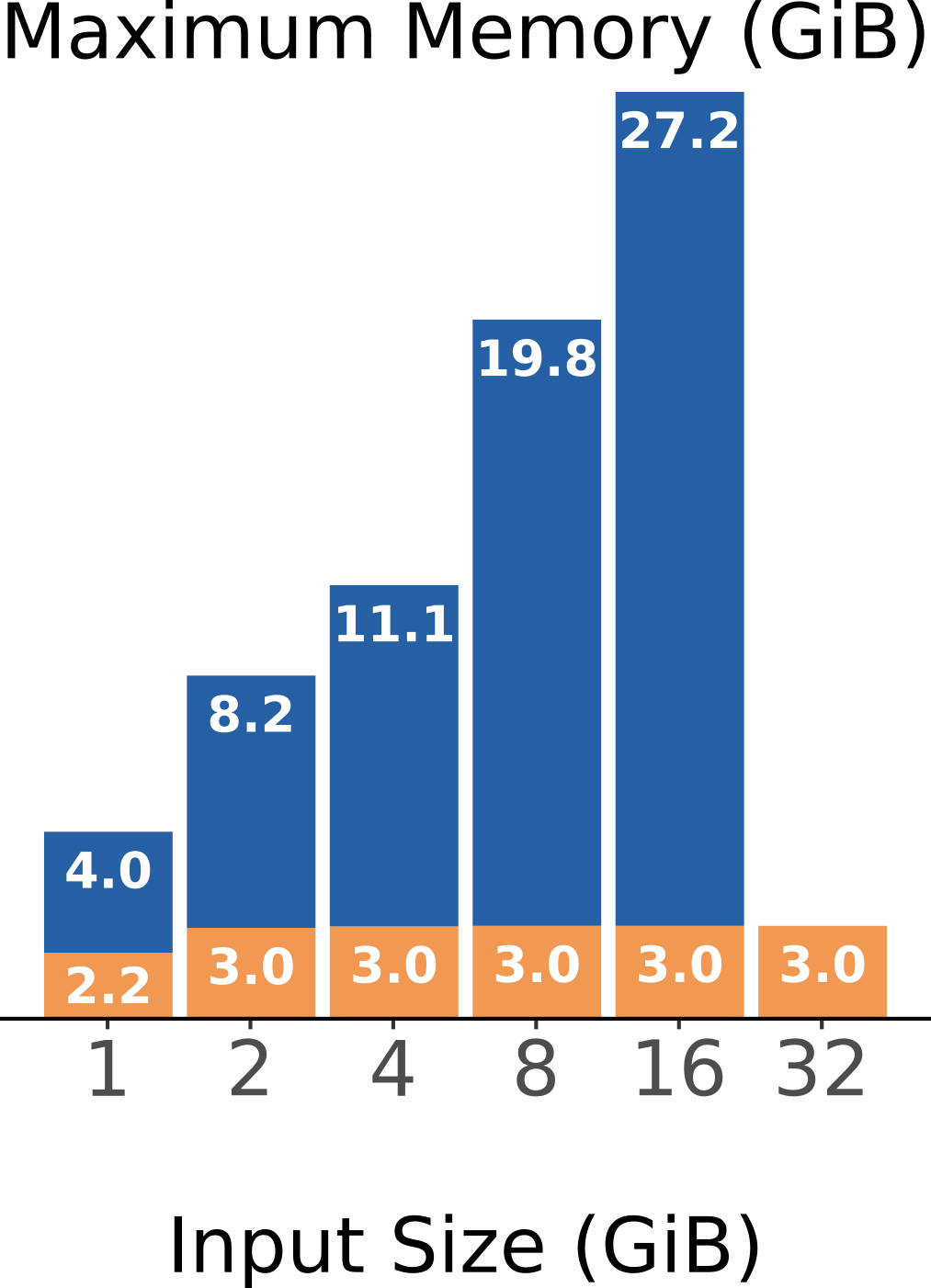}%
    \label{fig:res-max-mem}}
    \caption{Performance evaluation in terms of memory footprint.}
    \label{fig:res-memory-footprint}
\end{figure}

The memory footprint results are summarized in Fig.~\ref{fig:res-memory-footprint}. Notably, \anon{Annotat3D} exhibited superior memory efficiency. It maintained a stable residual memory footprint of approximately $0.6$GiB, and a nearly constant peak memory requirement of $\sim 3$~GiB, regardless of dataset size.%. A similar behavior was found in peak memory consumption where \anon{Annotat3d} maintained a nearly constant amount of memory requirements, $3$GiB.

The results demonstrate that \anon{Annotat3D} overcomes both performance and scalability limitations of existing tools. Its robust memory management allows segmentation of datasets up to $32$~GiB with interactive responsiveness. In comparison, scikit-image fails due to excessive execution time, and cuCIM fails to process volumes larger than $8$ and $16$~GiB, for grayscale and binary images, respectively.

%The results demonstrate that \anon{Annotat3D} effectively addresses both performance and scalability limitations of existing tools for large-scale volumetric data. Due to its robust memory management, the \anon{Annotat3D} allow users segmenting datasets up to $32$GiB with a reasonable responsiveness. In turn, the scikit-image fail due to operational limit found associated to time constraints and thus a lack of capability to processing data in a reasonable responsibility, which is paramount for a interactive workflow with the human-in-the-loop of processing. Finally the cuCIM library showed great results in terms of execution time, but fail to handle large datasets that exceed $8$GiB, with we consider operations that operates with grayscale volumes, and  $16$GiB, that operates with binary volumes.

\subsection{Discussion}

The experimental results, combined with a detailed analysis, highlight fundamental trade-offs between Harpia and existing GPU-based libraries, particularly cuCIM, in the context of large-scale volumetric processing.

A crucial architectural distinction lies in memory management. Harpia implements strict internal control of GPU memory allocation and deallocation, preventing direct user access to GPU-resident data. Although this approach may introduce additional CPU-GPU transfers during workflows construction, it offers significant advantages in memory safety and scalability. Specifically, the platform integrates a native chunked execution mechanism, which transparently partitions workloads into sub-volumes, enabling processing of datasets larger than GPU memory capacity. This ensures robust and reproducible processing of arbitrarily large images, even on hardware with limited memory.

In contrast, cuCIM provides direct GPU memory access, enabling efficient, low-overhead workflows entirely within GPU memory. However, this flexibility comes at the cost of scalability and stability: datasets must fit within GPU memory, and residual memory growth occurs after repeated operations, as documented in the CUDA memory pool behavior.~\footnote{\url{https://tinyurl.com/e392czcm}},\footnote{\url{https://tinyurl.com/379vv3ru}}
Although beneficial for isolated workloads, this retention leads to resource contention, out-of-memory errors, and instability in multi-user or production environments.

Although such limitations could in principle be mitigated with auxiliary Python-level tools, they introduce additional complexity. In contrast, the native support for chunked, memory-safe processing in Harpia provides an immediate, production-ready solution that meets the operational requirements of large-scale, multi-user scientific imaging workflows, independent of hardware limitations.

%In contrast, the cuCIM provides direct access to GPU-resident data, facilitating the construction of efficient and low-overhead processing entirety within the GPU memory space. However, this flexibility comes at the expense of memory stability and scalability. The absence of an internal mechanism for handling datasets that exceed GPU memory imposes strict size limitations in production operations. Moreover, the cuCIM exhibits residual memory growth with each subsequent operation, as confirmed by both empirical observations and the documented behavior of the underlying CUDA memory pool~\footnote{\url{https://tinyurl.com/e392czcm}}~\footnote{\url{https://tinyurl.com/ykyrnw75}}. This memory retention, while beneficial for performance in isolated workloads, poses significant challenges in a multi-user or shared GPU environments, where uncontrolled memory occupation may lead to resource contention, out-of-memory errors, and system instability which is critical in production-grade scientific infrastructure. While it is feasible that limitations of cuCIM could be mitigated through the development of auxiliary tools at Python level, the native support for scalable, memory-safe processing in \anon{Annotat3D} provides a immediate and production-ready solution tailored to the specific constraints and operational requirements of large-scale scientific imaging workflows.

% \begin{itemize}
%     \item Recap.
%     \item Summary of contributions.
%     \item Highlight collaborative annotation, web-based tool, streaming support.
%     \item Limitations and future work
% \end{itemize}

\section{Conclusion and Future Work}
\label{sec:conclusion}

This work introduced {Harpia}, a new CUDA-based processing library integrated into \anon{Annotat3D}, designed to address the challenges of interactive segmentation and processing of large volumetric datasets in scientific imaging workflows. Harpia implements strict memory control and native chunked execution, enabling scalable operation on datasets that exceed single-GPU memory capacity, while ensuring predictable resource usage suitable for shared HPC environments.

Benchmarking showed that Harpia outperforms established frameworks such as cuCIM and scikit-image in scalability, memory efficiency, and processing speed. Additionally, the web-based interface and integrated resource management make it well-suited for multi-user, remote-access environments typical of synchrotron and microscopy facilities.

As future work, we plan to extend Harpia to support multi-GPU and heterogeneous computing architectures, enabling even larger dataset processing and improved performance. Further integration of advanced and large visual models and server inference mechanisms is also envisioned, enhancing segmentation  accuracy and interactivity in complex scientific imaging tasks.
\section*{Acknowledgment}
We thank the Brazilian Ministry of Science, Technology, and Innovation (MCTI) for funding this work through the Brazilian Center for Research in Energy and Materials (CNPEM). We also thank the research groups from the MOGNO, CATERETÊ, and CARNAÚBA beamlines for their valuable suggestions during the development of this work.

% trigger a \newpage just before the given reference
% number - used to balance the columns on the last page
% adjust value as needed - may need to be readjusted if
% the document is modified later
% \IEEEtriggeratref{8}
% The "triggered" command can be changed if desired:
%\IEEEtriggercmd{\enlargethispage{-5in}}

% references section

% can use a bibliography generated by BibTeX as a .bbl file
% BibTeX documentation can be easily obtained at:
% http://mirror.ctan.org/biblio/bibtex/contrib/doc/
% The IEEEtran BibTeX style support page is at:
% http://www.michaelshell.org/tex/ieeetran/bibtex/
\bibliographystyle{IEEEtran}
% argument is your BibTeX string definitions and bibliography database(s)
\bibliography{example}
%
% <OR> manually copy in the resultant .bbl file
% set second argument of \begin to the number of references
% (used to reserve space for the reference number labels box)
%\begin{thebibliography}{1}
%
%\bibitem{IEEEhowto:kopka}
%H.~Kopka and P.~W. Daly, \emph{A Guide to \LaTeX}, 3rd~ed.\hskip 1em plus
%  0.5em minus 0.4em\relax Harlow, England: Addison-Wesley, 1999.

%\end{thebibliography}

% that's all folks
\end{document}